\pdfoutput=1

\documentclass[11pt]{article}

\usepackage[preprint]{acl}

\usepackage{times}
\usepackage{latexsym}

\usepackage[T1]{fontenc}

\usepackage[utf8]{inputenc}

\usepackage{microtype}

\usepackage{inconsolata}

\usepackage{graphicx}
\usepackage{booktabs}
\usepackage{url}

\title{Prune or Retrain: Optimizing the Vocabulary of Multilingual Models for Estonian}

\author{Aleksei Dorkin \and Taido Purason \and Kairit Sirts \\
        Institute of Computer Science \\
        University of Tartu \\
        \texttt{\{aleksei.dorkin, taido.purason, kairit.sirts\}@ut.ee} \\ }

\begin{document}
\maketitle

\begin{abstract}
Adapting multilingual language models to specific languages can enhance both their efficiency and performance. In this study, we explore how modifying the vocabulary of a multilingual encoder model to better suit the Estonian language affects its downstream performance on the Named Entity Recognition (NER) task. The motivations for adjusting the vocabulary are twofold: practical benefits affecting the computational cost, such as reducing the input sequence length and the model size, and performance enhancements by tailoring the vocabulary to the particular language. We evaluate the effectiveness of two vocabulary adaptation approaches---retraining the tokenizer and pruning unused tokens---and assess their impact on the model's performance, particularly after continual training.
While retraining the tokenizer degraded the performance of the NER task, suggesting that longer embedding tuning might be needed, we observed no negative effects on pruning.
\end{abstract}

\section{Introduction}

Adapting multilingual pretrained language models to specific languages can enhance both their efficiency and performance~\cite{kuratov2019adaptation,mroczkowski2021herbert}. The adaptation generally involves continuously training the full model on language-specific data. This approach can be expensive and unsuitable for less-represented languages.

In this study, we explore how modifying the vocabulary of a multilingual language model to better suit the Estonian language affects its downstream performance. Compared to the previous works \cite{gee-etal-2022-fast,csaki2024sambalingo,tejaswi2024exploring}, we focus on training newly initialized embeddings rather than the specific initialization approaches. The motivations for adjusting the vocabulary are twofold:

\paragraph{1) Practical Benefits:} A language-specific vocabulary can reduce the length of tokenized sequences, leading to more efficient training and fine-tuning. Meanwhile, the vocabulary of a multilingual model has to accommodate for all the languages it supports, which results in a significant amount of unused tokens in a monolingual use case. Consequently, adapting the vocabulary to a single language either by pruning the tokenizer or training a new one leads to the decrease in vocabulary size. In turn, decreasing the vocabulary size reduces the overall model size, which can improve computational efficiency.

\paragraph{2) Performance Enhancement:} Tailoring the vocabulary to a particular language may improve the model's ability to understand and process text in that language, potentially boosting performance on language-specific tasks.
\\

Our goal is to evaluate the effectiveness of two vocabulary adaptation approaches---retraining the tokenizer and pruning unused tokens---and assess their impact on encoder models' performance in the Estonian language, particularly after continual training, which is evaluated by fine-tuning on the named entity recognition task.
In the retraining approach, we train a new tokenizer on the Estonian National corpus (ENC),\footnote{\url{https://doi.org/10.15155/3-00-0000-0000-0000-08C04M}} and use the resulting vocabulary to replace/adapt the encoder-based multilingual DeBERTa model~\cite{he2023debertav3improvingdebertausing}. We first train the new embeddings with other model parameters frozen, and then continue training the full model with the masked language modeling loss. Finally, we fine-tune the model with new/adapted vocabulary on the Estonian NER dataset~\cite{sirts-2023-estonian} to evaluate the effect of vocabulary optimization. In the second approach, we simply prune the model's initial vocabulary to only keep the tokens that are present in the ENC, and experiment with continuing the training of either only the embeddings or all model parameters.

\section{Related Work}

Adapting a tokenizer to a new domain or language has been mainly done in two ways: modifying an existing tokenizer or training a new tokenizer on the domain data. The main focus of previous works has been on the embedding initialization methods for new or extended vocabulary, which is not needed in cases of vocabulary pruning.

The vocabulary pruning has been previously explored by \citet{abdaoui-etal-2020-load}. The main motivation was that mBERT~\cite{devlin-etal-2019-bert}, for instance, allocates more than 51\% of its parameters to the embeddings layer, yet only a fraction of the vocabulary is used for a single given language. Accordingly, the proposed approach is to create monolingual models from multilingual mBERT by shrinking the vocabulary of the original model. 
To select the tokens to keep for a given monolingual model, the authors collected token frequency statistics from the Wikipedia of the target language, and used these frequencies to filter out the tokens that appeared in less than 0.05\% lines.
As a result, the monolingual models retained up to 23.8\% of the original vocabulary.

Tokenizer extension for BPE models is usually done by first training a new tokenizer and then adding non-overlapping tokens to the existing tokenizer's vocabulary \cite[etc]{csaki2024sambalingo, gee-etal-2022-fast}.
\citet{csaki2024sambalingo} investigated extending an existing tokenizer and found that a correctly implemented vocabulary extension does not negatively affect downstream performance. 
\citet{tejaswi2024exploring} also studied the vocabulary extension of LLMs, finding that a larger extension requires more pre-training data for optimal results.

\citet{gee-etal-2022-fast} introduced a method for fast vocabulary transfer (FVT) to train a domain-specific tokenizer. The embeddings for tokens shared between the new and original tokenizers were copied. 
The embeddings for new tokens were obtained by averaging their sub-token embeddings from the original tokenizer. 
The model was then further pre-trained on in-domain data using the masked language modeling (MLM) loss before fine-tuning on target tasks.
\citet{dagan2024getting} applied the FVT \cite{gee-etal-2022-fast} for LLMs and found that the tokenizer choice impacts the effectiveness and downstream performance of LLMs. Specifically, they found that adapting the model to a new tokenizer requires tens of billions of tokens of retraining to outperform the original tokenizer. 
While our training ENC training corpus is far smaller, containing only few billions of tokens, we are experimenting with encoder models that are much smaller than LLMs.

\section{Methodology}

The overall methodology of optimizing a model's vocabulary to Estonian entails two main steps: 1) modifying the content of the vocabulary and adjusting the embeddings accordingly, and 2) continuing the training of the whole model to align it better with the new vocabulary.
As the base multilingual model we select mDeBERTa v3---the multilingual version of DeBERTa V3~\cite{he2023debertav3improvingdebertausing}---the model which is considered the SOTA encoder model at the time of the writing.

We compare two methods to modify the vocabulary of the mDeBERTa v3 multilingual language model for Estonian. The first method involves training a new tokenizer on the Estonian National Corpus (ENC), while in the second method we simply prune the model's original vocabulary to remove tokens that are not used in the tokenized ENC.

\paragraph{Retraining the Tokenizer}

We retained all the original settings (such as special tokens and pre- and post-processing steps) from the base mDeBERTa v3 model and retrained the underlying SentencePiece tokenizer.
For new tokens introduced by the retrained tokenizer, we initialized their embeddings using the mean of the original embedding matrix, while for tokens present in both the original and new tokenizers, we copied the existing embeddings. We also adjust the token to id mapping and resize the embedding matrix.

To align the newly initialized embedding vectors with the rest of the model, we first train the model on the training corpus with only the embeddings unfrozen using the  masked language modeling (MLM).

\paragraph{Tokenizer Pruning}

The pruning process starts with applying the existing tokenizer to the training data. Then we observed what tokens in the vocabulary never appear in our training data and removed them from the vocabulary. After that the token to id mapping was adjusted and the embedding matrix was rearranged and resized. Since no new tokens were added, we retained the original embeddings for the remaining tokens.

In our experiments, we use an approach similar to \citet{abdaoui-etal-2020-load} with two key differences. Firstly, we do not use a frequency threshold, but rather keep all the tokens that do appear in the language-specific data. This results in our model retaining approximately 67\% of the original vocabulary. Secondly, we employ a larger data source (that also includes Wikipedia)---the Estonian National Corpus. Both differences are aimed at maximizing the vocabulary coverage.

\paragraph{Continuous Training with LoRA}

To simulate continual training and assess the model's adaptability after vocabulary modification, 
we applied Low-Rank Adaptation (LoRA)~\cite{hu2021loralowrankadaptationlarge} training with MLM objective to both models.

\section{Experimental Setup}

\paragraph{Training Data}

For training the new tokenizer, and training and validation of the MLM objective, we employed the Estonian National Corpus (ENC). \footnote{\url{https://doi.org/10.15155/3-00-0000-0000-0000-08C04M}} The corpus contains approximately 16M documents with documents coming from different domains such as old and contemporary literature, academic texts, Wikipedia pages and discussions, as well crawled web pages. We performed light deduplication on the corpus resulting in ca 3.4B tokens and randomly split it into train, validation, and test, with both validation and test sets containing 1\% of the documents.

\paragraph{Models Developed}

In our experiments we employ the base version of mDeBERTa V3 as our base model, and apply the previously described approaches---tokenizer retraining and pruning---to it. For tokenizer retraining, we settle for 32K tokens in the final vocabulary, and train the new tokenizer using the train split of the ENC. Meanwhile, for pruning we collect the statistics on the appearance of tokens in the base model vocabulary in the ENC train split, then we remove all tokens that never appear in the data. This results in the vocabulary size of approximately 169K tokens. For both approaches we resize the embedding matrix and rearrange the corresponding vectors, while for tokenizer retraining we initialize the vectors that has not previously appeared in the base model vocabulary using the mean of the embedding matrix. The models were trained on the University High-Performance Cluster~\cite{https://doi.org/10.23673/ph6n-0144} using up to two A100 80GB GPUs.

\paragraph{Embedding training} 

For both approaches, we tuned the embeddings for a single epoch on sequences of 128 tokens in half-precision.
The number of devices, per device batch size, and gradient accumulation steps were configured so that the global batch size was 3092. The warm-up ratio was set to 0.05.

\paragraph{Continuous training with LoRA}
Most of the training parameters remain the same for LoRA continuous training, except for the learning rate which we set to be 1e-3. The LoRA itself was configured with a rank of 4 for the update matrices, using a scaling factor ($\alpha$) of 32. A dropout rate of 0.1 was applied to the LoRA layers to prevent overfitting. The adaptation was applied to the attention mechanism components (query, key, and value matrices) as well as the dense feed-forward layers. No bias parameters were updated during training.

\paragraph{Fine-tuning on NER} 
Intuitively, a downstream task where the model has to produce classification scores for individual tokes in the input is affected the most by the vocabulary modification. The most common type of such task is likely Named Entity Recognition (NER). For the Estonian language EstNER~\cite{sirts-2023-estonian} is the most comprehensive NER dataset. It contains 46K sentences annotated with 11 entity classes.
To assess the performance of the modified models, we fine-tuned models on EstNER for 50 epochs in half-precision with a global batch size of 64. For each model version we repeated the process three times and recorded the highest achieved F1 score in each run. We report the mean and the standard deviation over the three runs.

\begin{table}[h]
    \centering
    \small
    \setlength{\tabcolsep}{3.5pt}
    \label{tab:tokenizer_vocab_tokens}
    \begin{tabular}{lrrr}
        \toprule
        \textbf{Model} & \textbf{\# Params} & \textbf{Vocab} & \textbf{Tok per} \\
        & & \bf Size & \bf Word \\
        \midrule
        EstBERT                       & 124M &50K  & 1.82 \\
        XLM-RoBERTa base            & 278M & 250K & 2.04 \\
        TartuNLP Est-RoBERTa                    & 278M & 250K & 2.04 \\
        EMBEDDIA Est-RoBERTa                   & 116M & 40K  & 1.69 \\
        \midrule
        mDeBERTa base             & 279M & 250K & 2.23 \\
        mDeBERTa base Tuned      & 110M & 32K  & 1.75 \\
        mDeBERTa base Pruned     & 215M & 169K & 2.23 \\
        \bottomrule
    \end{tabular}
    \caption{Statistics on vocabulary size, number of parameters, and tokens per word (estimated on the validation split of the ENC) for related models.} 
    \label{tab:tokenizer_vocab_tokens}
\end{table}

\begin{table*}[ht]
    \centering
    \small
    \begin{tabular}{l cc}
        \toprule
        \textbf{Model} & \textbf{F1 Score (Mean $\pm$ Std)} & \textbf{MLM Accuracy} \\
        \midrule
        EstBERT & $75.72 \pm 0.19$ & -\\
        XLM-RoBERTa base& $ 80.66 \pm 0.37$ & - \\
        TartuNLP EstRoBERTa & $81.37 \pm 0.28$ & - \\
        Embeddia EstRoBERTa & $\textbf{83.77} \pm 0.24$ & - \\
        mDeBERTa-base & $80.96 \pm 0.19$ & - \\
        \midrule
        mDeBERTa-base $\rightarrow$ Tuned Embeddings & $76.40 \pm 0.23$ & 15.86 \\
        mDeBERTa-base $\rightarrow$ Tuned Embeddings $\rightarrow$ LoRA & $77.58 \pm 0.47$ & 29.74 \\
        \midrule
        mDeBERTa-base $\rightarrow$ Pruned & $80.62 \pm 0.12$ & -\\
        mDeBERTa-base $\rightarrow$ Pruned $\rightarrow$ Tuned Embeddings & $ 80.45 \pm 0.22$ & 25.84 \\
        mDeBERTa-base $\rightarrow$ Pruned $\rightarrow$ LoRA & $80.62 \pm 0.10$ & 38.42 \\
        \bottomrule
    \end{tabular}
    \caption{EstNER Evaluation F1 and ENC MLM Accuracy scores (excluding baseline models).}
    \label{tab:f1_scores}
\end{table*}

\paragraph{Evaluation Metrics}
We evaluate the tokenization efficiency by calculating the token per word ratio for different tokenizers. 
We measure the performance of the models on MLM objective using word prediction accuracy. To evaluate the downstream NER task we use the F1 score.

\section{Results}

In addition to the DeBERTa baseline, we also compare with various other models, including both Estonian-specific EstBERT \cite{tanvir2020estbert}, XLM-RoBERTa base \cite{conneau-etal-2020-unsupervised}, an Estonian-specific EstRoBERTa finetuned from the XLM-RoBERTa\footnote{\url{https://huggingface.co/tartuNLP/EstRoBERTa}} and another EstRoBERTa model trained from scratch.\footnote{\url{https://huggingface.co/EMBEDDIA/est-roberta}}
 
\paragraph{Tokenizer efficiency}
We first present the impact on the models' size and tokenizer efficiency in Table~\ref{tab:tokenizer_vocab_tokens}. 
We observe that adopting the smaller 32K language-specific vocabulary (mDeBERTa base Tuned) leads to approximately 60\% reduction in the number of parameters and 20\% reduction in tokens per word.
Meanwhile, simply pruning the vocabulary (mDeBERTa base Pruned) results in ca 23\% reduction in the number of parameters.

\paragraph{Tokenizer Optimization Results}

The models with optimized vocabulary MLM accuracy and the downstream NER task F1-scores are shown in Table~\ref{tab:f1_scores}. The top part shows the results for the baseline mDeBERTa base and the comparison models. The baseline mDEBERTa is in line with the multilingual XLM-RoBERTa, but little bit worse than Estonian-specific RoBERTa models.
The middle section of the Table~\ref{tab:f1_scores} shows the results on the models with newly created 32K vocabulary both only after the embedding tuning and then after training continuation with LoRA. While training continuation with LoRA substantially improves the MLM accuracy, replacing the tokenizer led to a substantial decrease in NER performance, with the model average F1 score being below both the baseline multilingual models and two out of three language specific models.
The bottom section of the Table~\ref{tab:f1_scores} shows the results for the models with vocabulary pruning. Again, continuation with the LoRA training improves the MLM accuracy, while the NER results are in the same range with the baseline.
The embedding tuning and LoRA training took approximately 120 GPU hours each with the pruned model taking longer due to the large vocabulary size.

\section{Discussion}

While replacing the vocabulary of the mDeBERTa model with a smaller Estonian-specific vocabulary led to more efficient input tokenization, the results on the downstream NER task suffered even after both embedding layer training and subsequent full model training with LoRA.
First, suboptimal embedding initialization approach likely played a role in the observed outcome.
Secondly, it is likely that a single epoch of embedding tuning was insufficient to match the performance of the base model. 
The subsequent LoRA MLM training resulted in slightly reducing the gap between the base model and the model with the retrained tokenizer, however it also remained insufficient to recover the original model's performance.
We presume that training for longer, both the embeddings and the LoRA parameters, would further reduce that gap.

In contrast, we observe that the vocabulary pruning has no observable negative effect on the downstream task. Meanwhile, tuning of the embeddings appears to have little to no effect on the downstream task, which suggests that such tuning is redundant in case of pruning.
Surprisingly and similarly to embeddings tuning, continued training with LoRA had no observable benefit for the pruned model, despite the gains in the MLM accuracy.

Finally, we observed that the relation between the MLM accuracy and F1 on NER is not transparent. While we acknowledge that MLM accuracy scores with different vocabularies are not directly comparable, the absence of the effect on the NER result in the presence of a notable improvement in the MLM accuracy in the pruned model is puzzling.

\section{Conclusion}

In this study, we explored two options for optimizing the vocabulary of a multi-lingual model for the Estonian language.
In summary, we found that replacing the tokenizer with a retrained language-specific version noticeably degrades model performance on the downstream NER task, and one epoch of embedding layer training on a 3.4B word corpus did not suffice to restore it.
While LoRA offers efficient way for further training continuation, a single epoch was insufficient to mitigate the negative impact of the tokenizer replacement.
On the other hand, pruning unused tokens proved to be an effective method to reduce vocabulary size without compromising performance.

\section*{Acknowledgments}

This research was supported by the Estonian Research Council Grant PSG721 and Estonian Language Technology Program Grant EKTB104.

\bibliography{custom}




\end{document}